\documentclass[runningheads]{llncs}
\usepackage[T1]{fontenc}

\usepackage{graphicx}
\usepackage{booktabs}
\usepackage{amsmath}
\usepackage{subcaption}
\usepackage{multirow}
\usepackage{array}
\usepackage{makecell}
\usepackage{amssymb}
\begin{document}

\title{Hierarchy-Aware and Anatomy-Guided Learning for Lung Ultrasound Video Classification}
\titlerunning{Anatomy-Guided Learning for LUS Video Classification}

\author{Alya Almsouti\inst{1}\orcidID{0009-0005-6349-7231} \and
Lotfi Mecharbat\inst{1} \and
Noha Aboukhater\inst{2} \and
Yousef Alabrach\inst{3} \and
Siddiq Anwar\inst{4} \and
Andre Kumar\inst{5} \and
Ibrahim Almakky\inst{1} \and
Mohammad Yaqub\inst{1}}

\authorrunning{A. Almsouti et al.}

\institute{
Mohamed bin Zayed University of Artificial Intelligence, UAE\\
\email{alya.almsouti@mbzuai.ac.ae}
\and
SEHA, UAE
\and
Sheikh Khalifa Medical City, UAE
\and
Khalifa University, UAE
\and
Stanford University, USA
}

\maketitle
\begin{abstract}
% Lung ultrasound (LUS) is a practical bedside tool for assessing pulmonary congestion, particularly in renal-care populations at risk of fluid overload. 
Lung ultrasound (LUS) is a bedside tool for assessing pulmonary edema in patients at risk due to heart failure or impaired kidney function. However, automated LUS analysis remains challenging because of speckle noise, imaging artifacts, and operator-dependent acquisition variability. In this work, we present a deep learning framework for multi-class LUS video classification that explores two components: hierarchy-aware training, and anatomy-guided learning. Starting from a strong baseline, we introduce hierarchical training strategies and then introduce pleural line mask supervision to guide model attention toward anatomically relevant regions. We study four clinically relevant classes--healthy, B-lines, consolidations, and mixed B-lines with consolidations--using an open-access dataset of 1,886 videos from 219 patients, evaluated with patient-level five-fold cross-validation. Results show that hierarchy-aware training improves pathological separation relative to flat classification, while mask-guided attention supervision achieves the highest mean macro-F1 of 65.7\% and produces more localized attention patterns. Transfer experiments on the external COVID-BLUeS dataset further show competitive and parameter-efficient adaptation while preserving pleural-focused attention behavior. These findings suggest that combining clinically structured objectives with anatomy-guided supervision is a practical approach to robust, interpretable LUS video analysis. Code and model implementations are available at \url{https://github.com/Alya-Almsouti/LUS-video-classification}.
\keywords{Lung ultrasound \and Multi-class classification \and Hierarchical learning \and Attention supervision \and Pulmonary edema}
\end{abstract}
\section{Introduction}

Pulmonary edema is the accumulation of excess fluid in the lung interstitium and alveolar spaces, which impairs gas exchange. It is a frequent and clinically significant complication in patients with heart failure, acute kidney injury (AKI), chronic kidney disease (CKD), and dialysis-dependent renal failure. Importantly, imaging signs of pulmonary congestion can appear before overt respiratory symptoms, making early detection essential for timely intervention and optimized fluid and treatment management \cite{aki_pulmonary_edema,aki_ckd_pulmonary_edema,early_detection_imp_2}.
In clinical settings, pulmonary edema is typically assessed using physical examination, lung auscultation, chest radiography, serum biomarkers, and bedside ultrasound; however, while these tools are complementary, ultrasound is the most sensitive modality for early extravascular lung water \cite{pocovidnet,joseph2022covecho,maw2019diagnostic,subramanyam2023multi}.

\begin{figure}[t]
    \centering
    
    \begin{subfigure}[b]{0.3\textwidth}
        \centering
        \includegraphics[width=\textwidth]{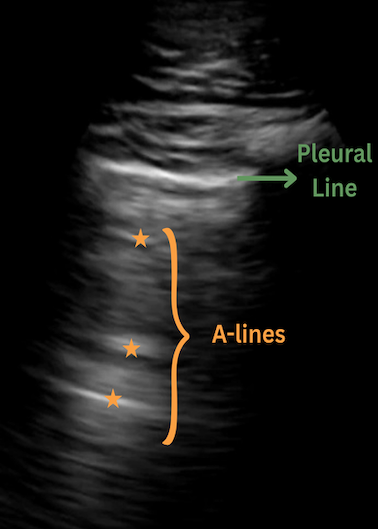}
        \caption{}
        \label{fig:lungus:alines}
    \end{subfigure}
    \begin{subfigure}[b]{0.3\textwidth}
        \centering
        \includegraphics[width=\textwidth]{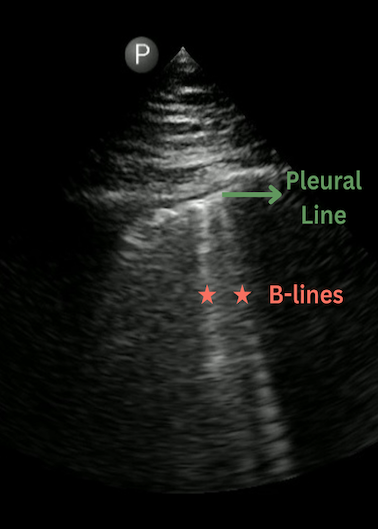}
        \caption{}
        \label{fig:lungus:blines}
    \end{subfigure}
    \begin{subfigure}[b]{0.3\textwidth}
        \centering
        \includegraphics[width=\textwidth]{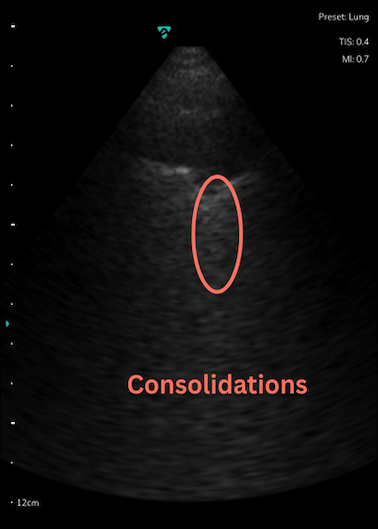}
        \caption{}
        \label{fig:lungus:consolidations}
    \end{subfigure}
    
    \caption{Lung ultrasound artifacts: (a) A-lines and pleural line; (b) B-lines and pleural line; (c) Consolidations.}
    \label{fig:lungus:artifacts}
\end{figure}

Lung ultrasound (LUS), widely used within point-of-care ultrasound (POCUS), provides a bedside, radiation-free method for real-time assessment of pulmonary congestion, allowing for repeated evaluations as clinical conditions evolve \cite{joseph2022covecho,pocovidnet,maw2019diagnostic}. This makes it particularly well-suited to both acute and longitudinal care settings, where dynamic fluid management decisions are required.
Key findings in LUS are centered on pleural line appearance and artifact patterns. In a normally aerated lung, an intact pleural surface generates repeated horizontal reverberation artifacts (A-lines; Fig. \ref{fig:lungus:alines}). As lung fluid increases and aeration decreases, vertical comet-tail artifacts (B-lines; Fig. \ref{fig:lungus:blines}) emerge from the pleural line and extend to the bottom of the scan. With more advanced pathology, when fluid accumulates within the alveoli, subpleural tissue-like patterns appear (consolidations), often accompanied by pleural irregularity (Fig.~\ref{fig:lungus:consolidations}) \cite{pleura_hmm_2020,aline_morph_2021,consolidation_on_lus}.
Despite its clinical utility, LUS interpretation remains operator-dependent and is affected by ultrasound-specific noise, acquisition variability, and reader experience, making robust interpretation challenging for novice sonographers in busy workflows. This motivates automated LUS analysis as both a clinical decision-support tool and an educational aid.
The main contributions of this work are:
\begin{itemize}
    \item A LUS-specific framework that adapts hierarchy-aware objectives and mask-guided attention supervision for video classification.
    
    \item A SAM2-based semi-automatic video annotation pipeline that substantially reduced labeling time and enabled curation of a clinician-annotated pleural line segmentation dataset.

    \item Improved classification and interpretability: best multi-class performance, with attention maps more consistently focused on pleural and subpleural regions.

    \item Transfer experiments on the external COVID-BLUeS severity and COVID detection tasks show competitive, parameter-efficient transfer while preserving pleural-focused attention behavior.
\end{itemize}

\section{Related Work}

\subsubsection{Classical Machine Learning and Model-Based Methods.}
Early approaches for lung ultrasound (LUS) relied on handcrafted features and explicit geometric priors, with pleural line detection as a key task \cite{pleura_hmm_2020,chen2021automated,aline_morph_2021}. Representative pipelines used HMM-based modeling with Viterbi tracking, morphological and adaptive filtering for A-line enhancement, and Radon-transform-based line detection. While these approaches offered a degree of interpretability, their performance was often degraded by speckle noise, probe-angle variability, and cross-device domain shifts, ultimately limiting their generalization and scalability.

\subsubsection{Supervised Deep Learning Methods.}
During the COVID-19 pandemic, LUS research shifted toward supervised deep learning for diagnosis, localization, and severity assessment. Early CNN-based approaches, such as those by Roy et al. and POCOVIDNet, leveraged class activation maps and confidence-based heuristics for post hoc interpretation, but offered limited spatial precision and weak anatomical grounding \cite{pocovidnet,roy2020deep}. 
Subsequent work moved beyond binary COVID-19 classification toward artifact localization, segmentation, and multi-class analysis. Detection-based models and U-Net variants were increasingly used to localize and segment B-lines and pleural patterns \cite{joseph2022covecho,subramanyam2023multi}. However, many of these studies relied on private datasets, which constrained reproducibility \cite{abbasi2025improved,howell2024deep,munoz2024deep,wiedemann2025covid}.

\subsubsection{Representation Learning.} More recent work has targeted the domain gap between natural and ultrasound images through ultrasound-specific representation learning. HiCo, USFM, spatio-temporal contrastive learning, and ultrasound foundation models improved transferability \cite{chen2023contrastive,chen2021uscl,jiao2024usfm,zhang2022hico}. However, key limitations remain for pulmonary edema assessment: many studies still emphasize frame-level predictions, ignoring important temporal information, and rely on private datasets with limited reproducibility. In addition, many end-to-end systems still provide limited anatomy-guided interpretability, with insufficient explicit supervision toward pleural line regions.

\subsubsection{Attention Supervision and Hierarchical Learning.}
Mask- or localization-guided supervision has been used to encourage classifiers to attend to clinically relevant regions, including ViT-based attention supervision methods such as LGM-ViT and SPAN \cite{lgmvit,span}. Hierarchical classification has also been used to exploit clinically meaningful label structure, such as normal-vs-abnormal separation followed by subtype classification \cite{11067957}. Our work adapts these ideas to video-level LUS classification, using pleural line masks as anatomical supervision for multi-class pulmonary edema assessment.

\subsubsection{Rationale for Hierarchical and Anatomy-Guided Learning.}
The clinical decision process is naturally hierarchical: separating healthy from pathological findings is generally easier than distinguishing among pathological subtypes. This motivates hierarchy-aware objectives that encode the diagnostic structure during training. At the same time, clinicians visually anchor interpretation around the pleural line and adjacent subpleural region, where
B-lines originate and consolidation patterns often emerge. This motivates anatomy-guided learning that explicitly guides model attention toward clinically relevant regions. 
% Thus, the novelty of this work lies in the LUS-specific integration and evaluation of these ideas, together with the curation of pleural-line masks for attention supervision in video-level pulmonary edema classification.

\section{Methodology}

Given a video consisting of $N$ frames, we denote the input sequence as
$\mathbf{X} = (\mathbf{x}_1, \ldots, \mathbf{x}_N)$, where each frame
$\mathbf{x}_t \in \mathbb{R}^{3 \times H \times W} ,\quad t = 1, \ldots, N $ represents an RGB ultrasound
image.
Each frame is processed independently by a frame encoder $E(\cdot)$, which maps
the input image to a $d$-dimensional latent representation
$\quad \mathbf{z}_t = E(\mathbf{x}_t)$, $\mathbf{z}_t \in \mathbb{R}^{d}$.
The resulting sequence of frame-level embeddings is denoted as
$\mathbf{Z} = (\mathbf{z}_1, \ldots, \mathbf{z}_N) \in \mathbb{R}^{N \times d}$.
To obtain one video-level representation, we apply a temporal
aggregation function $T(\cdot)$ over the sequence
$\mathbf{v} = T(\mathbf{Z}).$

Finally, the aggregated video-level representation is passed to a classifier
head to produce the predicted logits $\hat{\mathbf{y}} = g(\mathbf{v})$. For classification, the training objective is defined as cross-entropy $\mathcal{L}_{\text{cls}}$ between predicted logits and the ground-truth label $y$.

\subsection{Hierarchical Training}
Because separating healthy from abnormal cases is clinically easier than distinguishing among abnormal subtypes, we encode this diagnostic structure directly in the training objective.
\subsubsection{Hard Hierarchical Classification.}
In the hard hierarchical setting, separate binary models are trained for
different sub-tasks and combined only at inference time. We evaluate two
variants.

\paragraph{Two-stage hierarchical strategy.}
In the first strategy, two independent classifiers are trained: healthy vs.
B-lines and healthy vs.
consolidations. This yields
\begin{equation}
\hat{y}_{\text{HB}} = g_{\text{HB}}(T(E(\mathbf{X}))), \qquad
\hat{y}_{\text{HC}} = g_{\text{HC}}(T(E(\mathbf{X}))).
\end{equation}
At inference, the final label is determined from the two outputs: both healthy
$\rightarrow$ healthy; only one pathological output $\rightarrow$ that pathology;
both pathological outputs $\rightarrow$ B-lines + consolidations.

\paragraph{One-versus-all strategy.}
The second strategy uses one-versus-all training with three classifiers:
healthy-vs-not-healthy, B-lines-vs-not-B-lines, and
consolidations-vs-not-consolidations, yielding
$\hat{y}_{\text{H}}, \hat{y}_{\text{B}}, \hat{y}_{\text{C}}$. During inference,
if $\hat{y}_{\text{H}}$ is healthy, the final label is healthy; otherwise,
$\hat{y}_{\text{B}}$ and $\hat{y}_{\text{C}}$ determine the pathology label. If
both pathological classifiers are negative, healthy is assigned as fallback.

\subsubsection{Soft Hierarchical Training with Multiple Heads.}
In the soft hierarchical setting, a single shared model is trained with two
task-specific heads. The temporal aggregator produces two video representations:
$\mathbf{v}_{\text{healthy}}$ and $\mathbf{v}_{\text{path}}$, which are processed by
$g_{\text{healthy}}(\cdot)$ and $g_{\text{path}}(\cdot)$, respectively.
The first head predicts healthy vs.
pathological and the second head predicts the pathology subtype:
\begin{equation}
\hat{\mathbf{y}}_{\text{healthy}} = g_{\text{healthy}}(\mathbf{v}_{\text{healthy}}), \quad
\hat{\mathbf{y}}_{\text{path}} = g_{\text{path}}(\mathbf{v}_{\text{path}})
\end{equation}
where $\hat{\mathbf{y}}_{\text{healthy}} \in \mathbb{R}^{2}$ and $\hat{\mathbf{y}}_{\text{path}} \in \mathbb{R}^{3}$.
% denotes logits for healthy and pathological classes ,corresponds to logits for B-lines, 
% , and B-lines + consolidations.

Training uses a hierarchical loss. Let
$y_{\text{healthy}} \in \{0,1\}$ denote healthy ($0$) vs.
pathological ($1$). Additionally, let  $y_{\text{path}}$ be the ground-truth pathological class. The binary loss and subtypes loss are: 

\begin{equation}
\mathcal{L}_{\text{healthy}} =
\mathrm{CE}(\hat{\mathbf{y}}_{\text{healthy}}, y_{\text{healthy}}), \quad
\mathcal{L}_{\text{path}} =
\mathrm{CE}(\hat{\mathbf{y}}_{\text{path}}, y_{\text{path}}).
\end{equation}

The total training loss is then defined as
\begin{equation}
\mathcal{L}(y) =
\begin{cases}
\lambda_{\text{healthy}} \, \mathcal{L}_{\text{healthy}}, &
\text{if } y_{\text{healthy}} = 0, \\[6pt]
\lambda_{\text{healthy}} \, \mathcal{L}_{\text{healthy}}
+ \lambda_{\text{path}} \, \mathcal{L}_{\text{path}}, &
\text{if } y_{\text{healthy}} = 1,
\end{cases}
\end{equation}
where $\lambda_{\text{healthy}}$ and $\lambda_{\text{path}}$ are scalar weights
that control the relative contribution of the hierarchical loss terms.

\subsection{Mask-Guided Attention Supervision}
\begin{figure}[t]
    \centering
    \includegraphics[width=\linewidth]{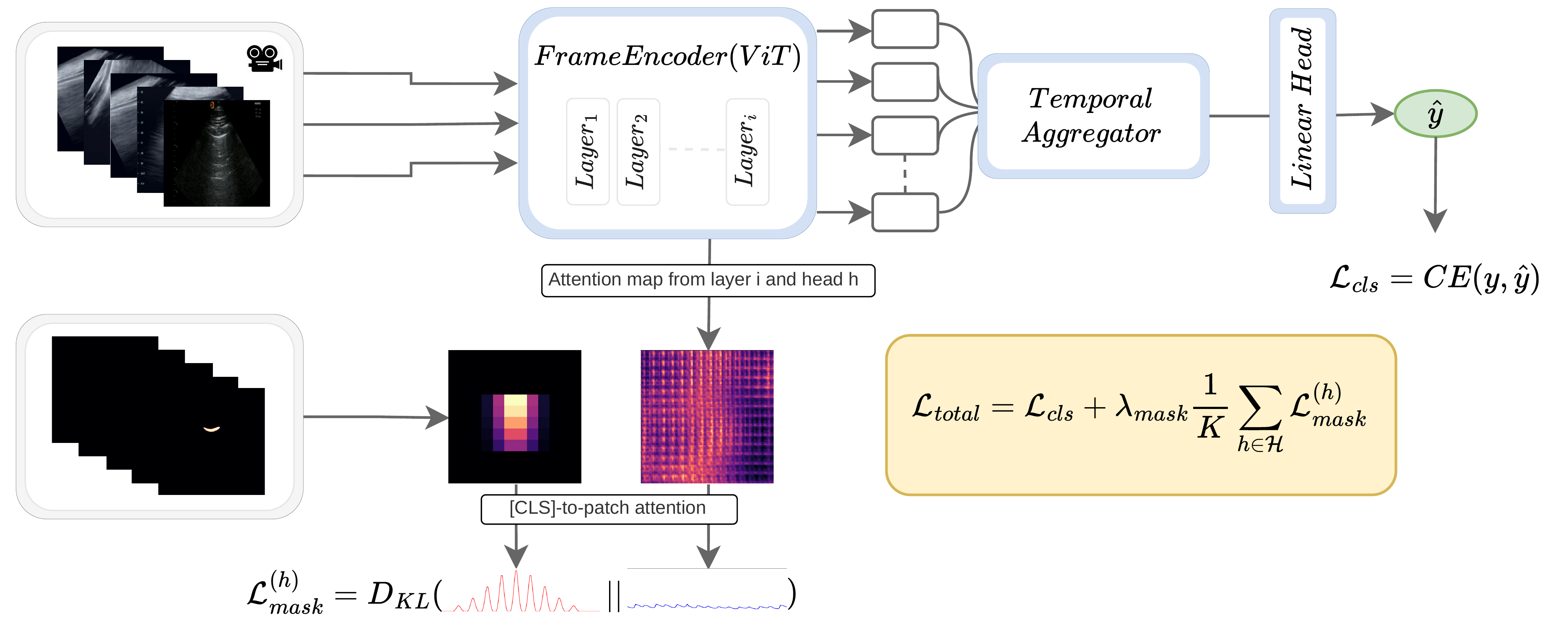}
    \caption{\textbf{Mask-Guided Attention Supervision.} Video frames are first encoded by a ViT backbone and aggregated temporally for final classification, while pleural line masks are converted into target \texttt{[CLS]}-to-patch attention distributions to supervise selected attention heads toward anatomically relevant pleural regions.}
    \label{fig:anatomy_aware_diagram}
\end{figure}

Because clinicians primarily inspect the pleural line and adjacent subpleural region when interpreting LUS, we explicitly guide part of the model attention toward these anatomically relevant areas. This approach is illustrated in Fig. \ref{fig:anatomy_aware_diagram}.
We keep the same classification architecture and add only an attention
supervision term to selected ViT heads of a certain block. Let
$\mathbf{A} \in \mathbb{R}^{H \times N_{\text{tok}} \times N_{\text{tok}}}$
be the self-attention tensor of a chosen transformer block, with
$N_{\text{tok}} = 1 + L$ (one \texttt{[CLS]} token and $L$ patch tokens) and $H$ is the number of heads. Supervision
is applied to a head subset $\mathcal{H} \subset \{1,\dots,H\}$, where
$|\mathcal{H}| = K$.

Given a binary pleural line mask
$\mathbf{m} \in \{0,1\}^{1 \times 224 \times 224}$,
we downsample it to patch resolution 
$\mathbf{m}^{p} =
\text{AvgPool}_{16}(\mathbf{m})
\in
\mathbb{R}^{1 \times 14 \times 14}.$
We then apply light smoothing, use progressive dilation across heads — where dilation is applied once for every two heads to encourage different heads to focus on neighborhoods of varying size around the pleural line —
and convert each mask into a normalized patch-level target distribution constructing a target \texttt{[CLS]}-to-patch attention distribution:
\begin{equation}
\mathbf{q}
=
\text{Normalize}\big(
\text{Flatten}(\mathbf{m}^{p})
\big),
\qquad
\mathbf{q} \in [0,1]^{L}.
\end{equation}
For each supervised head $h \in \mathcal{H}$, we extract
\texttt{[CLS]}-to-patch attention: $\mathbf{A}^{(h)}_{\text{cls}}
=
\mathbf{A}[h,0,1:]
\in
\mathbb{R}^{L}.$
Mask supervision is imposed using KL divergence:
\begin{equation}
\mathcal{L}_{\text{mask}}^{(h)}
=
\mathrm{D_{KL}}\!\left(
\mathbf{A}^{(h)}_{\text{cls}}
\,\|\, 
\mathbf{q}
\right).
\end{equation}
The final mask loss averages over supervised heads 
$\mathcal{L}_{\text{mask}}
=
\frac{1}{K}
\sum_{h \in \mathcal{H}}
\mathcal{L}_{\text{mask}}^{(h)}.$ 
Finally, the overall objective augments classification loss:
\begin{equation}
\mathcal{L}_{\text{total}} =
\mathcal{L}_{\text{cls}} +
\lambda_{\text{mask}} \, \mathcal{L}_{\text{mask}},
\end{equation}
where $\lambda_{\text{mask}}$ controls the strength of the attention supervision. We evaluated Block~12 as a late-stage supervision point, where patch features are expected to be more semantically mature, and Block~8 as an intermediate point to test whether earlier anatomical guidance can shape attention before the final layers. We compared 3 and 6 supervised heads to examine partial versus stronger attention constraints.

\section{Experimental Design}
\subsection{Datasets and Segmentation Dataset Curation}
We use the open-access Stanford LUS dataset \cite{kumar2025openaccess},
containing 1,886 videos from 219 patients (up to 12 videos per patient,
depending on lung zones). Each video is labeled as one of four classes:
healthy, B-lines, consolidations, or B-lines + consolidations.
To further assess the transferability and generalizability of our method, we evaluated on the COVID-BLUeS dataset \cite{wiedemann2025covid}. Because COVID-BLUeS does not provide the same four-class label space as the Stanford LUS dataset, we used the dataset-defined severity prediction and COVID detection tasks. The first task predicts severity of POCUS lung videos on a four-level scale (0--3), ranging from healthy (0) to severe (3), while the second task performs binary COVID detection (COVID vs. non-COVID).

Pleural line annotations were not provided in the open-access LUS dataset; therefore, a semi-automatic pipeline was developed and open-sourced \cite{videosam2026} to obtain a segmentation dataset for mask-guided attention supervision.
Initially, four medical doctors annotated the videos by splitting the dataset, such that each video was annotated once. Using the annotation platform, annotators placed point prompts on a selected frame from each video. Positive points were used to indicate pleural line regions, while negative points marked non-pleural areas. These prompts were provided to SAM2 \cite{ravi2024sam} to generate segmentation masks, which were iteratively refined by the annotators based on the model output. Once satisfactory masks were obtained, they were aggregated across videos and manually cleaned to form an initial pleural line segmentation dataset.
This produced 49,084 masks (15\% of total frames) from 329 videos and 43 patients. Next, a
U-Net model was trained on these curated masks and 
used for iterative
auto-labeling and refinement (two cycles). The final segmentation set contained
147,342 frames from 941 videos and 200 patients (45\% of total frames). Although no independent pleural line ground truth was available for Dice/IoU evaluation, the generated masks underwent manual quality control and cleaning to ensure accurate coverage of the pleural line. Since the masks are downsampled to ($14\times14$) patch resolution before constructing the attention target, the supervision relies on accurate localization of the pleural line rather than pixel-perfect boundary delineation. This semi-automatic pipeline substantially reduced labeling time by replacing dense, frame-wise manual segmentation with sparse point prompting on selected frames.

\subsection{Evaluation and Training Setup}
Evaluation used patient-level five-fold cross-validation with precomputed splits.
Each video was represented as a fixed-length sequence of $N=150$ frames. For videos exceeding this length, frames were sampled according to a Gaussian distribution over the temporal axis centered at the midpoint, and videos with fewer than $N$ frames were padded. All selected frames were then resized to $224 \times 224$. Additional details on the choice of fixed sequence length and temporal sampling strategy are provided in the supplementary material.

During training, we applied data augmentation to improve robustness, including
random resized cropping (scale: 0.7--1.0), random horizontal flipping, and
random rotations up to $10^{\circ}$. To better match ultrasound acquisition
conditions, speckle noise was randomly injected into training
samples. All images were normalized using the ImageNet mean and standard
deviation. Validation data underwent only resizing and normalization.

Training ran for up to 100 epochs with a 3-epoch probing stage (frame encoder
frozen), followed by end-to-end fine-tuning. AdamW optimizer (weight decay 0.01) was used
with cosine annealing. Learning rates were $1\times10^{-5}$ for non-encoder
parameters and $5\times10^{-6}$ for encoder parameters.
Models were trained with cross-entropy, label smoothing (0.1), and fold-specific
balanced class weights. Validation was monitored per epoch using macro-F1,
weighted-F1, and balanced accuracy. Early stopping used
macro-F1 (patience 20), and the best macro-F1 checkpoint per fold was reported.
Unless noted otherwise, all results are validation-set metrics summarized over
the five folds.

\subsection{Baselines}
In our baseline experiments, we varied both the frame encoder $E(\cdot)$ and the
temporal aggregation function $T(\cdot)$ to identify which backbone and
temporal aggregation strategy were most effective for the ultrasound modality
and video-level representation learning.
We compared four representative frame encoders: (i) an Ultrasound Foundation
Model (USFM) \cite{jiao2024usfm}, (ii) HiCo \cite{zhang2022hico},
(iii) ResNet-18 \cite{he2016deep}, and (iv) ViT-Small
\cite{dosovitskiy2021image}. For all variants, only the backbone image encoder
was used, and any task-specific classification heads were removed.
USFM is a self-supervised ultrasound foundation model trained on a large dataset consisting of 2,187,915 images from different organs including the lung \cite{jiao2024usfm}. HiCo is a self-supervised
ultrasound representation learning framework trained on 23,231 images extracted
from lung and liver ultrasound videos \cite{zhang2022hico}. 

Given frame-level embeddings
$\mathbf{Z} = (\mathbf{z}_1, \ldots, \mathbf{z}_N)$, we evaluated four
video-level aggregation strategies:
(i) \textbf{Majority Voting (MV)}, where each frame was classified independently and
the final video-level prediction corresponded to the most frequent class among
frame-level predictions; during training, the video label was applied to each
frame;
(ii) \textbf{Mean Pooling (MP)}, where temporal information was aggregated as
\begin{equation}
\mathbf{v} = \frac{1}{N}\sum_{t=1}^{N} \mathbf{z}_t;
\end{equation}
(iii) \textbf{Weighted Pooling (WP)}, where normalized attention weights
$\{\alpha_t\}_{t=1}^{N}$ were computed from the frame embeddings to form
\begin{equation}
\mathbf{v} = \sum_{t=1}^{N} \alpha_t \mathbf{z}_t,
\qquad \text{with } \sum_{t=1}^{N} \alpha_t = 1;
\end{equation}
(iv) \textbf{Transformer-Based Aggregation (T)}, where a temporal transformer was applied
to $(\mathbf{z}_1, \ldots, \mathbf{z}_N)$ and forms the video representation by
concatenating the output $[\mathrm{CLS}]$ token with the mean-pooled
representation of the remaining tokens for final video-level classification.

\begin{table}[htbp]
\centering
{\setlength{\tabcolsep}{3pt}
\caption{Comparison of hierarchy-aware learning and anatomy-guided learning variants for four-class classification. Results are reported as macro-F1 (mean $\pm$ standard deviation) across folds. Class 0 indicates healthy class, Class 1: B-lines, Class 2: Consolidations, Class 3: B-lines + Consolidations.}
\label{tab:baseline_hierarchical_mask}
\begin{tabular}{lccccc}
\hline
\textbf{Model-} & \multicolumn{5}{c}{\textbf{F1-score}} \\
\cline{2-6}
\textbf{Configuration} & \textbf{Class 0} & \textbf{Class 1} & \textbf{Class 2} & \textbf{Class 3} & \textbf{Macro} \\
\hline
% \multicolumn{6}{l}{\textit{Baselines}} \\
USFM-MV & $79.8\pm 8.40$ & $50.0\pm10.1$ & $25.6\pm12.0$ & $37.8\pm9.00$ & $49.8\pm6.70$ \\
USFM-MP & $76.3\pm 6.30$ & $53.1\pm6.60$ & $7.7\pm7.80$ & $31.6\pm9.50$ & $43.5\pm7.20$ \\
HiCo-MV & $81.2\pm 6.00$ & $57.5\pm5.30$ & $37.0\pm7.70$ & $41.6\pm15.0$ & $54.1\pm2.10$ \\
ResNet18-T & $89.2\pm 4.60$ & $56.9\pm3.60$ & $35.9\pm2.20$ & $41.5\pm14.1$ & $56.1\pm4.00$ \\
ViT-Small-T & $91.6\pm 3.50$ & $67.6\pm5.50$ & $51.8\pm6.00$ & $40.1\pm9.50$ & $62.8\pm3.20$ \\
\hline

\multicolumn{6}{l}{\textit{Hierarchy-aware - Hard hierarchy}:} \\

\quad Two-Stage & \textbf{98.4$\pm$1.10} & $39.2\pm8.10$ & $10.1\pm7.10$ & $38.4\pm6.20$ & $46.5\pm2.20$ \\
\quad One-vs-all  & \underline{92.4 $\pm$ 3.10} & \underline{71.3 $\pm$ 7.30} & $40.0\pm9.30$ & $\textbf{52.2$\pm$5.80}$ & $64.5\pm1.60$ \\

\multicolumn{6}{l}{\textit{Hierarchy-aware - Soft hierarchy} ($\lambda_{\text{path}} = 1$):} \\
\quad $\lambda_{\text{healthy}} = 1$ & $87.2\pm7.50$ & $63.3\pm8.70$ & $41.1\pm3.30$ & $46.0\pm10.3$ & $59.4\pm3.00$ \\
\quad $\lambda_{\text{healthy}} = 2$ & $89.5\pm5.20$ & $68.6\pm6.00$ & $51.0\pm4.50$ & $43.0\pm9.10$ & $63.4\pm3.00$ \\
\hline
\multicolumn{6}{l}{\textit{Mask-guided attention supervision} ($\lambda_{\text{mask}} = 2$):} \\
Block 8, 3 Heads & $92.1\pm4.50$ & $70.4\pm7.90$& \textbf{53.0$\pm$7.2}& $45.5\pm8.30$ & \textbf{65.7$\pm$2.40} \\
Block 12, 3 Heads  & $92.2\pm3.80$ & \textbf{71.4$\pm$6.60} & $48.3\pm6.00$& \underline{49.0 $\pm$ 7.00} & \underline{65.2 $\pm$ 1.10} \\
Block 12, 6 Heads  & $90.2\pm6.00$& $68.2\pm7.4$& $45.6\pm6.20$& $47.1\pm10.1$ & $62.8\pm2.70$ \\
\hline
\shortstack{\textit{Hierarchy+}\\\textit{anatomy-guided}}  & $90.4\pm5.00$& $69.0\pm7.40$& $51.0\pm1.4$& $44.2\pm12.9$& $64.0\pm2.50$ \\

\end{tabular}}
\end{table}

\section{Results and Discussion}

\label{section:baseline_results}
Results obtained using different frame encoders and temporal aggregation methods are presented in Table \ref{tab:baseline_hierarchical_mask}.
To improve readability, the table reports the main configurations used for comparison, while the full evaluation of all backbones across all temporal aggregation strategies is provided in the supplementary material.
Two key observations emerge. First, the choice of frame encoder had a greater impact on performance than the temporal aggregation method. USFM consistently performed the worst across all aggregation methods, followed by HiCo. ResNet18 showed variable performance, while ViT-Small consistently achieved the best results under all settings.
This highlights that the quality of latent embeddings is a key factor in determining video classification model performance. Notably, although USFM and HiCo were pretrained on ultrasound scans (including lung), the ViT model initialized with ImageNet weights demonstrated superior transferability for lung pathology discrimination.
This may reflect a mismatch between ultrasound pretraining and the target task: USFM learns broad multi-organ ultrasound representations, while HiCo was pretrained on a smaller lung/liver dataset, whereas ViT-Small may provide more general visual features that adapt effectively during fine-tuning.

Second, the class-wise F1 scores were consistent across all backbone and aggregation configurations. The Healthy class (Class 0) achieved the highest performance, followed by B-lines (Class 1), while the lowest performance was observed for Consolidations and the mixed class (B-lines + Consolidations; Classes 2 and 3). 
This indicates that the lower performance for these classes may be due to the encoder learning representations of healthy patterns more effectively than those of pathological cases. In particular, extracting sufficiently discriminative latent embeddings for consolidation and mixed patterns appears to have been more challenging, possibly due in part to their lower representation across folds, as shown by the fold-level class distributions in the supplementary material.
Moreover, the choice of backbone and temporal aggregator substantially influenced per-class performance. The most notable improvement was observed for Consolidations (Class 2), with an absolute gain of 44 \% (from 7.7\% to 51.8\%). Overall, ViT-Small with transformer-based aggregation was selected as the default frame encoder and temporal aggregator for the rest of the experiments.

\subsection{Effect of Hierarchy-Aware Learning}
The hierarchical experiments in Table \ref{tab:baseline_hierarchical_mask} (\textit{Hierarchy-aware}) showed that the way hierarchy was enforced significantly affected performance. The two-stage strategy achieved the highest score for the healthy class (98.4\%), likely because it relied on two independent classifiers and predicted healthy only when both agreed, resulting in a stricter and more confident decision boundary. In contrast, performance on B-lines and consolidations was lower, as the models were trained separately against the healthy class without explicitly learning to distinguish between pathologies. This lack of inter-pathology discrimination led to confusion between abnormal classes, reflecting the limitation of optimizing primarily for healthy vs. abnormal separation.
On the other hand, the one-versus-all hierarchy achieved the strongest overall hierarchical performance (64.5\%). This approach improved Class~1 and substantially boosted Class~3 performance by 12.1\%, the highest gain observed across all methods, while maintaining strong healthy detection (92.4\%). Compared to the two-stage design, this resulted in a more balanced class-wise performance. These findings suggest that explicitly training each classifier against all alternatives provided a stronger discriminative signal, leading to improved separation between pathological subtypes, particularly for the B-lines class.
The observed performance for the soft hierarchy was as follows. With equal weighting ($\lambda_{\text{healthy}} = 1,\ \lambda_{\text{path}} = 1$), performance remained close to the baseline, with no substantial gains. Increasing the loss weight on the healthy class ($\lambda_{\text{healthy}} = 2$) led to an overall improvement, yielding modest gains over both the equal-weight setting and the baseline. Although the one-versus-all strategy achieved higher performance across hierarchical settings, it required training three separate classifiers, whereas the soft hierarchy relied on a single model, highlighting a trade-off between performance and model complexity.

\subsection{Effect of Mask-Guided Attention Supervision}
\label{section:mask_guided_results}

The mask-guided attention supervision results (Table~\ref{tab:baseline_hierarchical_mask}-\textit{Mask-guided attention supervision}) showed that enforcing \texttt{[CLS]} attention toward anatomically relevant regions consistently improved performance across configurations. The most notable gains were observed when supervision was applied at Block 8 with 3 heads (heads 0, 1, 2), yielding the best overall performance (65.7\%) and the highest improvement for Class~2, suggesting that guiding mid-level representations toward the pleural line was particularly effective for capturing pathology-specific patterns.
Applying supervision at Block 12 also improved performance, with the 3-head setting achieving strong results for Class~1 and Class~3, indicating that later-layer refinement further enhanced class-specific discrimination. However, increasing the number of supervised heads to 6 (all heads) led to a drop in performance, suggesting that over-constraining attention may limit the model’s ability to capture complementary cues beyond the pleural line region.

Overall, both Block~8 and Block~12 supervision improved over the ViT-Small-T baseline, with Block~8 using 3 supervised heads achieving the highest mean macro-F1. The drop observed when supervising all 6 heads suggests that partial head supervision is preferable, likely because it guides attention toward pleural regions while preserving other heads for complementary visual cues. This analysis should be interpreted as a targeted ablation of selected supervision locations and head counts, rather than an exhaustive layer-wise and head-wise search.

Because fold-level variance was non-negligible, we further performed paired fold-level comparisons for the main configurations. The mask-guided model improved over the ViT-Small-T baseline on all five folds and achieved the highest mean macro-F1, but the difference did not reach conventional statistical significance with $n=5$ folds (Wilcoxon signed-rank test, $p=0.0625$). Therefore, we interpret the gain as a favorable performance trend rather than definitive statistical superiority. Additional confidence intervals, paired tests, and aggregated confusion matrices are provided in the supplementary material.

\begin{figure}[ht!]
  \centering
  \begin{subfigure}[t]{\textwidth}
    \includegraphics[width=\linewidth]{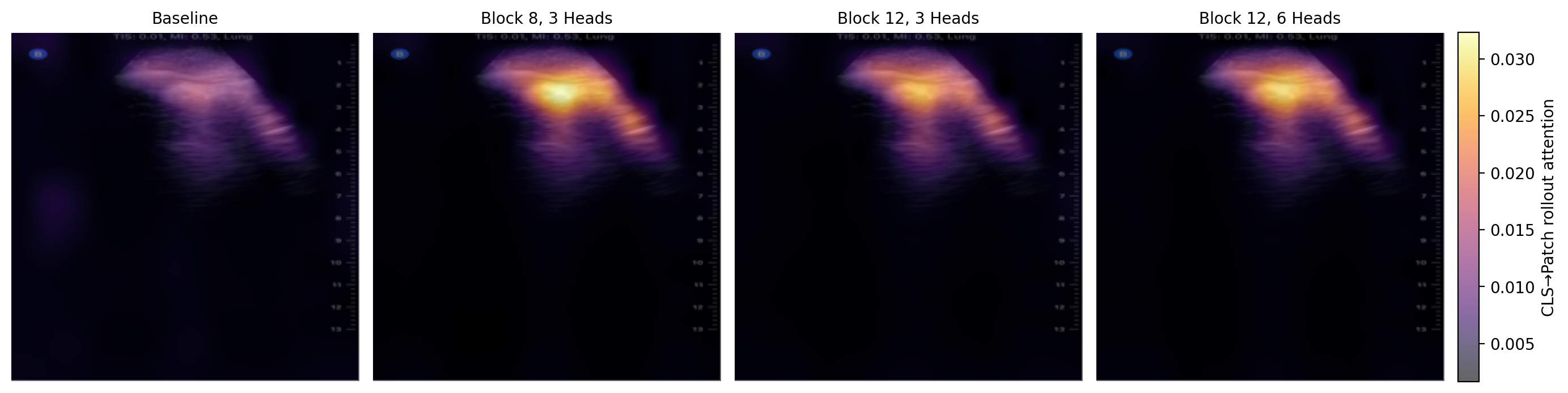}
  \end{subfigure}\hfill
  \begin{subfigure}[t]{\textwidth}
    \includegraphics[width=\linewidth]{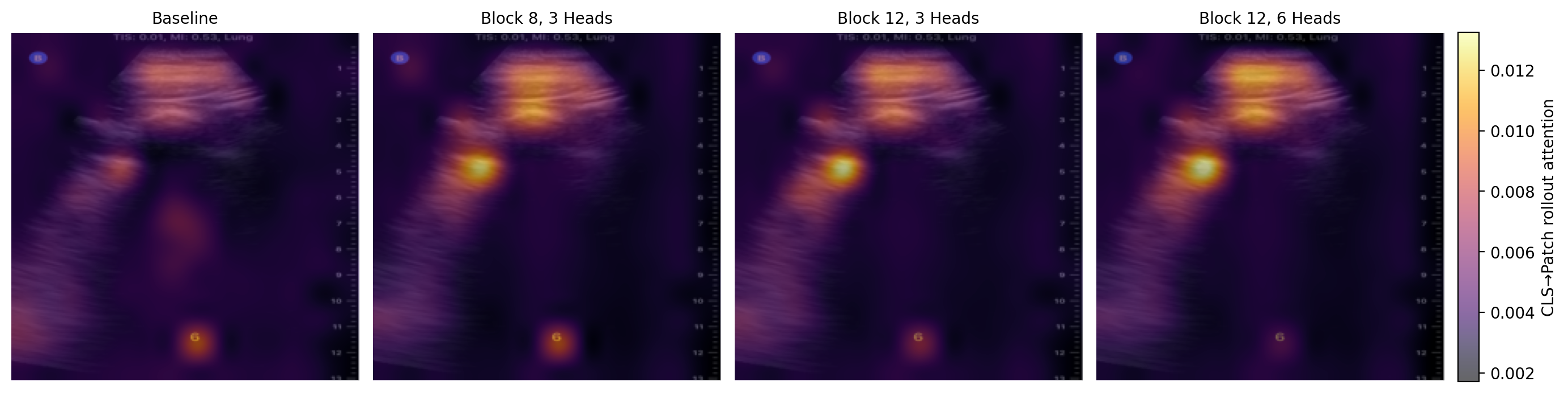}
  \end{subfigure}\hfill
  \begin{subfigure}[t]{\textwidth}
    \includegraphics[width=\linewidth]{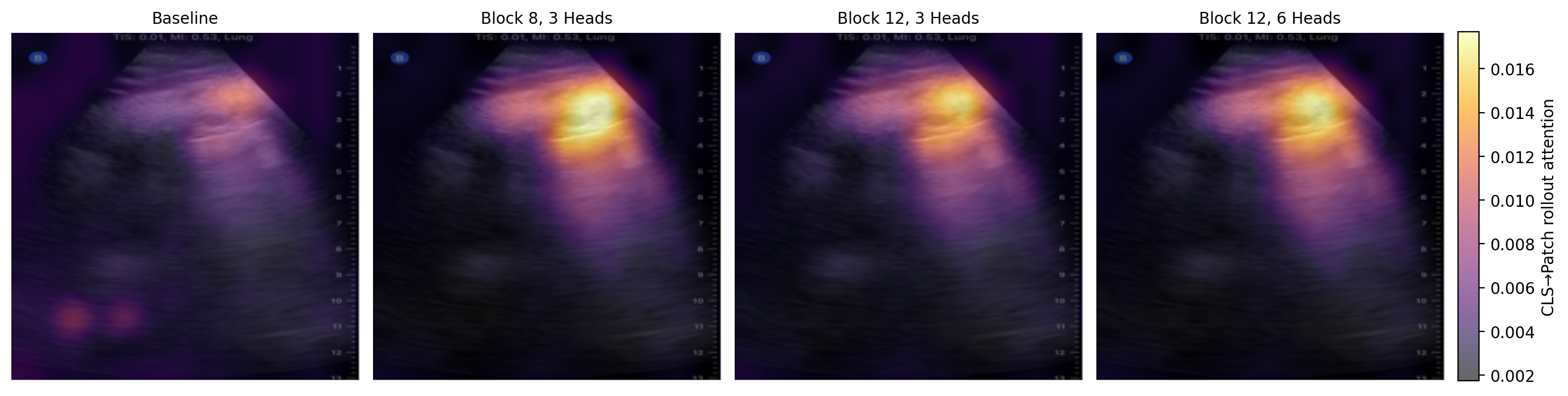}
  \end{subfigure}

  \caption{Comparison of CLS-to-patch attention rollout maps across model configurations (as described in Section \ref{section:mask_guided_results}). Each column shows the spatial attention distribution over the input image, highlighting regions that contribute most strongly to the final frame embedding.}
  \label{fig:maskloss_heatmaps}

\end{figure}

The effect of mask-guided supervision is illustrated in Fig.~\ref{fig:maskloss_heatmaps}, which presents qualitative \texttt{[CLS]}-to-patch attention rollouts. The baseline model exhibits relatively diffuse attention, while mask-supervised models show clearer concentration near pleural structures. Taken together, anatomy-guided learning improved localization and yielded modest but consistent classification gains, whereas combining hierarchical learning and mask-guided attention supervision did not provide further improvement over the best anatomy-guided-only setting. One possible explanation is that the hierarchical objective and anatomy-guided loss impose partially overlapping constraints on the shared representation; therefore, when explicit pleural line attention supervision is already applied, the additional hierarchy loss may reduce optimization flexibility rather than provide complementary gains.
An additional qualitative example is provided in the supplementary material.

\begin{table}[t]
\centering
\caption{Transferability to the COVID-BLUeS severity task. Results are reported on the four-class severity setting (0--3), mean across the 5 folds.}
\label{tab:severity_transferability}
\begin{tabular}{>{\raggedright\arraybackslash}p{5.6cm}ccccc}
\hline
\textbf{Model} & \textbf{Accuracy} & \textbf{F1} & \textbf{Precision} & \textbf{Recall} & \makecell{\textbf{Trainable}\\\textbf{Params}} \\
\hline
USFM-Majority Voting & 31.5 & 25.3 & 31.5 & 23.0 & 86.14M \\
HiCo-Majority Voting & 30.9 & 25.3 & 30.9 & 22.8 & 561.64M \\
ResNet-18-Transformer & 45.5 & 43.7 &  45.5 & 46.0 & 15.73M\\
ViT-Small-Transformer & \textbf{46.6} & \underline{46.0} & \textbf{46.6} & 49.4 & 26.15M \\
\multicolumn{5}{l}{\textit{Transfer settings from pretrained model (previous task)}} \\
Linear head only (new head) & 44.0 & 40.7 & 44.0 & 47.1 & 0.01M \\
Temporal aggregator + head fine-tuned & 42.1 & 40.2 & 42.1 & 47.4 & 4.41M \\
Full fine-tuning & \textbf{46.6} & \textbf{46.1} & \textbf{46.6} & \underline{48.2} & 26.15M  \\
\hline
\end{tabular}
\end{table}

\begin{table}[t]
\centering
\caption{Transferability to the COVID-BLUeS COVID detection task. Results are mean across the 5 folds.}
\label{tab:covid_transferability}
\begin{tabular}{>{\raggedright\arraybackslash}p{5.6cm}ccccc}
\hline
\textbf{Model} & \textbf{Accuracy} & \textbf{F1} & \textbf{Precision} & \textbf{Recall} &\makecell{\textbf{Trainable}\\\textbf{Params}} \\
\hline
ViT-Small-Transformer & 61.4 & 72.7 & \textbf{94.3} & 59.2 & 26.15M \\
\multicolumn{5}{l}{\textit{Transfer settings from pretrained model (previous task)}} \\
Linear head only (new head) & 66.1 & 70.2 & 74.2 & \textbf{68.0} & 0.01M \\
Temporal aggregator + head fine-tuned & \textbf{66.2} & \textbf{72.9} & 83.3 & 65.6 & 4.41M \\
Full fine-tuning & 62.0 & 71.7 & 88.9 & 62.0 & 26.15M  \\
\hline
\end{tabular}
\end{table}

% \begin{figure}[t]
%   \centering
%     \includegraphics[width=\linewidth]{Figures/grid_blocks_heads.png}
%   \caption{Mean attention distance per head across transformer encoder blocks for four model configurations (as described in Section \ref{section:mask_guided_results}). Each subplot corresponds to one model (see subplot title), and each curve represents an encoder block. The plotted value is the mean spatial distance (in pixels) between the pleural line mask-center token and all tokens, weighted by the CLS attention distribution (CLS $\rightarrow$ $j$), and averaged over the Fold 0 validation set. Shaded regions indicate $\pm 1$ standard deviation across samples.}
%   \label{fig:maskloss_lineplot}
% \end{figure}

\subsection{Transferability to External LUS Severity and COVID Tasks}

To evaluate the transferability of the anatomy-guided model, we conducted experiments on an external dataset across two tasks.

For the severity prediction task (Table \ref{tab:severity_transferability}), all frame encoders were paired with their best-performing temporal aggregators identified from the baseline experiments. In addition, the best-performing anatomy-guided model was fine-tuned under different transfer settings to assess its adaptability.
The results show that ultrasound-pretrained models (USFM and HiCo) performed poorly in this setting, despite their large capacity, suggesting limited task alignment. In contrast, models trained from scratch on the target dataset (ResNet-18 and ViT-Small) and full fine-tuning of the pretrained anatomy-guided model achieved competitive performance, indicating that leveraging the full model capacity and adapting all parameters was beneficial for this task.
Notably, partial transfer strategies remained effective. Fine-tuning only the temporal aggregator or training a new classification head achieved reasonably competitive results while using significantly fewer trainable parameters. This highlights a clear trade-off between performance and efficiency, where lightweight adaptation offers a practical alternative when computational resources are constrained. 

For COVID detection (Table \ref{tab:covid_transferability}), a different transfer pattern was observed. Temporal-aggregator+head fine-tuning achieved the best overall performance (F1 $72.9$, accuracy $66.2$) while updating only $4.41$M parameters, outperforming both full fine-tuning (F1 $71.7$, accuracy $62.0$, $26.15$M parameters) and the directly trained ViT-Small baseline (F1 $72.7$, accuracy $61.4$). Overall performance differences across methods were modest, but more parameter-efficient strategies consistently matched or exceeded full fine-tuning.

Our results on COVID-BLUeS are not directly comparable to those reported in the original study, as we could not identify the exact cross-validation split files from the released material. Therefore, we generated new patient-level splits, with all videos from a given patient kept within the same fold. For context, the original study reported best patient-level F1 scores of 81\% for COVID detection and 51\% for severity prediction. Thus, our experiments should be interpreted as an external transfer evaluation on COVID-BLUeS tasks rather than a strict reproduction of the original benchmark.

\begin{figure}[t]
  \centering

  % --- COVID task ---
  \textbf{(a) COVID Task}\\[0.5em]

  \begin{subfigure}[t]{\textwidth}
    \centering
    \includegraphics[width=\linewidth]{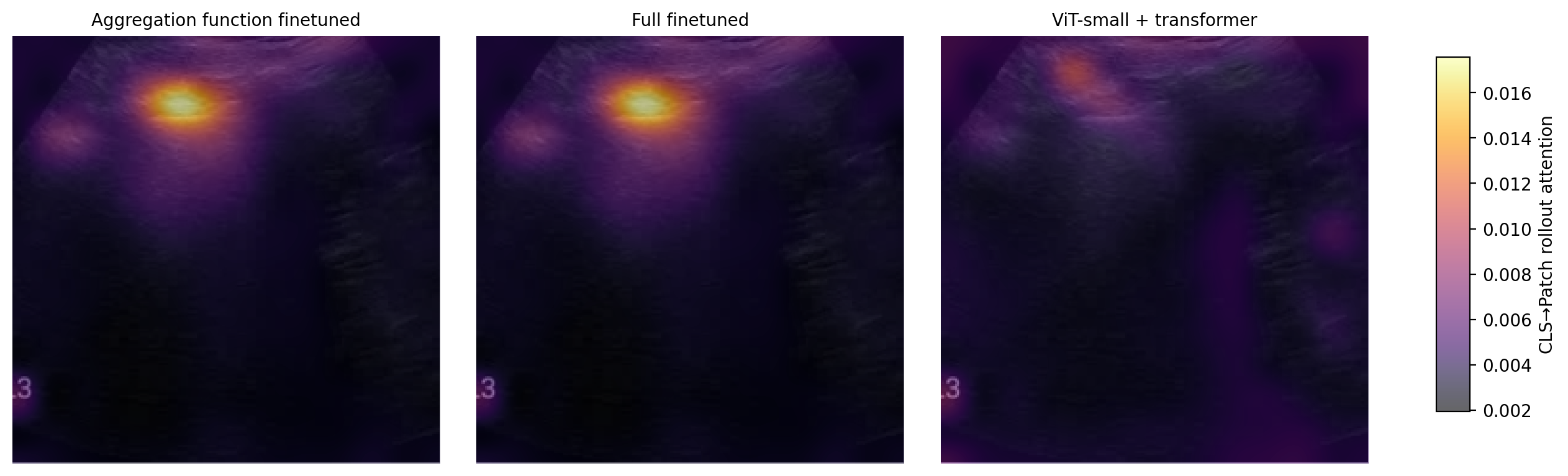}
  \end{subfigure}\hfill
  \begin{subfigure}[t]{\textwidth}
    \centering
    \includegraphics[width=\linewidth]{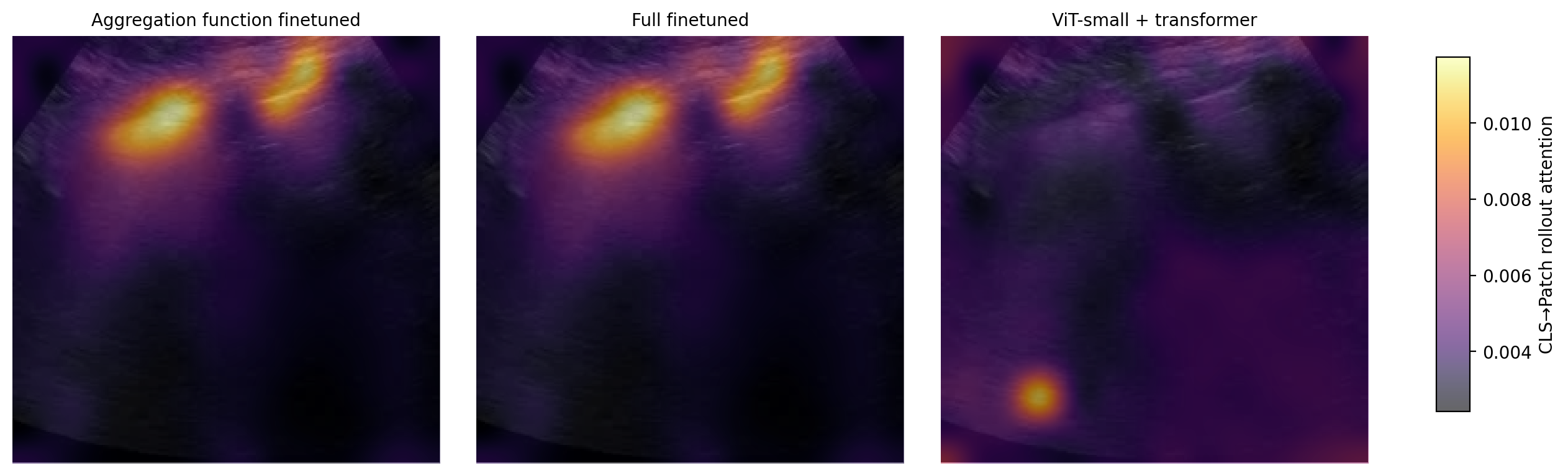}
  \end{subfigure}

  \vspace{1em}

  % --- Severity task ---
  \textbf{(b) Severity Task}\\[0.5em]

  \begin{subfigure}[t]{\textwidth}
    \includegraphics[width=\linewidth]{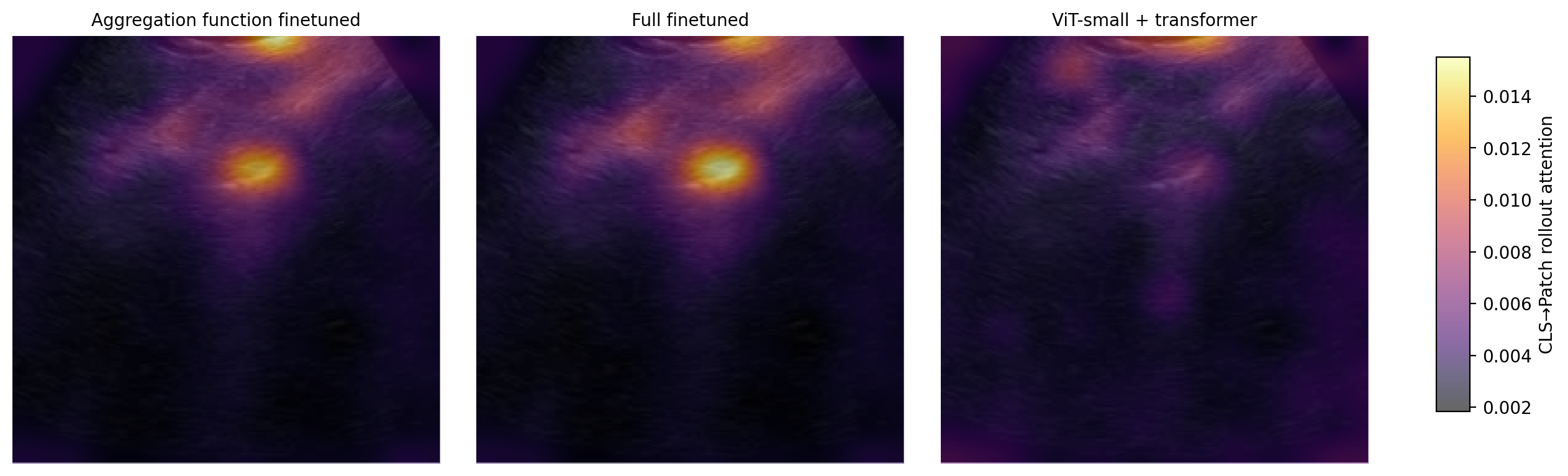}
  \end{subfigure}
  \begin{subfigure}[t]{\textwidth}
    \includegraphics[width=\linewidth]{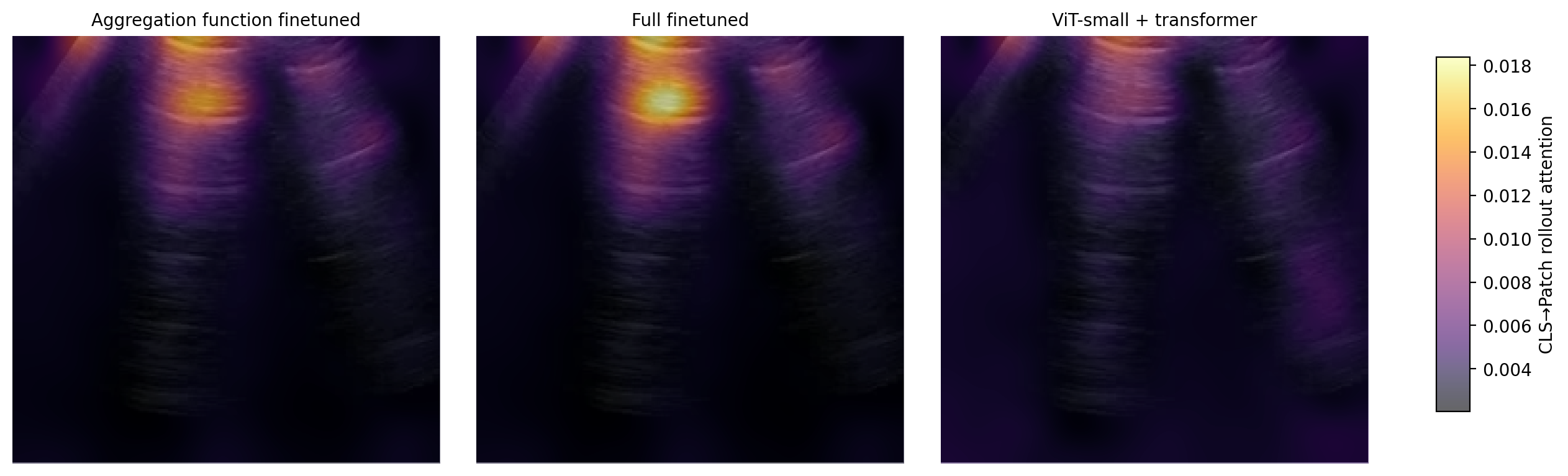}
  \end{subfigure}

  \caption{Comparison of CLS-to-patch attention rollout maps across models fine-tuned on the COVID-BLUeS dataset. Each column visualizes the spatial attention distribution over the input image, highlighting regions that contribute most to the final frame embedding.}
  \label{fig:heatmaps_covid_blues}
\end{figure}

Qualitatively (Fig.~\ref{fig:heatmaps_covid_blues}), the directly trained ViT-Small Transformer baseline attended to regions outside clinically relevant areas, indicating less stable anatomical focus. In contrast, attention rollouts from transfer-based models (temporal-aggregator+head fine-tuning and full fine-tuning) remained consistently concentrated around the pleural line and adjacent subpleural region, even without mask-guided supervision during transfer. This suggests that the pretrained backbone preserved anatomically meaningful attention patterns and effectively generalized them to the external dataset.

% \subsection{Calibration and Predictive Reliability}

% Beyond discrimination performance, we assessed calibration to evaluate reliability of predicted probabilities for clinical decision support. We compared representative baseline, hierarchy-aware, and anatomy-guided models using standard calibration metrics (e.g., ECE, Brier score, and NLL), where lower values indicate better calibration.

% Preliminary trends suggest that improvements in macro-F1 do not always translate to improved calibration, highlighting a common accuracy--reliability trade-off in deep classifiers. This motivates reporting calibration alongside classification metrics in LUS settings where confidence estimates may influence downstream decisions. Overall, the best-performing model should therefore be selected based on both discriminative performance and probability reliability.

\section{Limitations}
This study is limited by the moderate patient cohort size, class imbalance, and possible label ambiguity, particularly for consolidation and mixed-pathology cases, which were consistently harder to discriminate. During pleural line annotation, clinicians were allowed to flag label disagreements, and 34 of 329 videos (10.4\%) were relabeled, suggesting that some cases contained ambiguous pathology patterns. Pleural line masks were generated semi-automatically and manually quality-checked; however, independent dense ground-truth masks and multi-annotator agreement were not available. Although attention supervision improved pleural-region localization, formal clinician evaluation of the attention maps remains future work.

\section{Conclusion}
This paper presents a LUS video classification framework built on clinically meaningful problem framing, hierarchy-aware training, and mask-guided attention supervision. Under patient-level evaluation on a public Stanford dataset, hierarchical objectives improve multi-class discrimination, and mask-guided attention supervision achieves the highest mean macro-F1 while improving anatomical localization. The results suggest that clinically structured objectives and explicit spatial guidance are practical directions for robust and interpretable pulmonary edema assessment from POCUS videos.

\clearpage
\bibliographystyle{splncs04}
\bibliography{references}

\end{document}